\documentclass[a4paper]{article}
\usepackage{graphicx} 
\usepackage[utf8]{inputenc}
\usepackage{geometry}
\usepackage{caption}
\usepackage{subcaption}
\usepackage{multirow}
\usepackage{hyperref}

\title{TMComposites: Plug-and-Play Collaboration Between\\Specialized Tsetlin Machines\thanks{The code for building TM Composites can be found at \href{https://github.com/cair/Plug-and-Play-Collaboration-Between-Specialized-Tsetlin-Machines}{https://github.com/cair/Plug-and-Play-Collaboration-Between-Specialized-Tsetlin-Machines}.}}
\author{Ole-Christoffer Granmo\\Centre for AI Research\\University of Agder\\Norway\\\texttt{ole.granmo@uia.no}}
\date{}

\begin{document}

\maketitle

\begin{abstract}
Tsetlin Machines (TMs) provide a fundamental shift from arithmetic-based to logic-based machine learning. Supporting convolution, they deal successfully with image classification datasets like MNIST, Fashion-MNIST, and CIFAR-2. However, the TM struggles with getting state-of-the-art performance on CIFAR-10 and CIFAR-100, representing more complex tasks. This paper introduces plug-and-play collaboration between specialized TMs, referred to as \emph{TM Composites}. The collaboration relies on a TM's ability to specialize during learning and to assess its competence during inference. When teaming up, the most confident TMs make the decisions, relieving the uncertain ones. In this manner, a TM Composite becomes more competent than its members, benefiting from their specializations. The collaboration is plug-and-play in that members can be combined in any way, at any time, without fine-tuning. We implement three TM specializations in our empirical evaluation: Histogram of Gradients, Adaptive Gaussian Thresholding, and
Color Thermometers. The resulting TM Composite increases accuracy on Fashion-MNIST by two percentage points, CIFAR-10 by twelve points, and CIFAR-100 by nine points, yielding new state-of-the-art results for TMs. Overall, we envision that TM Composites will enable an ultra-low energy and transparent alternative to state-of-the-art deep learning on more tasks and datasets.
\end{abstract}

\section{Introduction}

The power of state-of-the-art machine learning comes from encoding enormous amounts of historical data using millions (and lately, trillions) of parameters. However, the historical data makes the trained models carry on biases, discrimination, and prejudices~\cite{Bender2021Parrots}, while the number of parameters makes them incomprehensible to humans~\cite{Rudin2019}.

Researchers are further accumulating evidence that models based on correlation are brittle. Even state-of-the-art deep learning models, with their high computational cost and carbon footprint~\cite{Schwartz2020GreenA}, tend to learn simple correlations instead of capturing the underlying causal dynamics of the data~\cite{scholkopf2021causal,sauer2021counterfactual}. Relying on correlations rather than causation is problematic when correlations are spurious or accurate only in limited contexts due to data bias.

Recently, the emerging paradigm of TMs~\cite{granmo2018tsetlin} has made a fundamental shift from arithmetic-based to logic-based machine learning. Seen from a logical engineering perspective~\cite{lucas1995}, a TM produces propositional/relational clauses in Horn form (AND rules)~\cite{Saha2022}. However, the TM is data-driven, learning the clauses by employing efficient finite state machines, so-called Tsetlin automata~\cite{Tsetlin1961}. By using multiple clauses to signify confidence, TMs handle uncertainty despite its logic-based origin~\cite{abeyrathna2020confidence}. In this way, TMs introduce the concept of logically interpretable learning, where both the learned model and the learning process are easy to follow and explain.

While the TM supports convolution~\cite{granmo2019convtsetlin} and has dealt successfully with MNIST, Fashion-MNIST, and CIFAR-2~\cite{maheshwari2023}, it struggles with getting state-of-the-art performance on CIFAR-10 and CIFAR-100~\cite{dropclause}. In this paper, we propose one possible approach to this challenge by introducing a \emph{team} of specialized TMs, enabling collaboration. The paper contributions are the following:
\begin{itemize}
    \item We first propose a novel architecture for plug-and-play collaboration between specialized TMs. The architecture is plug-and-play because independently pre-trained TMs can be connected at any time and in any combination without fine-tuning. Their individual confidences are simply normalized and aggregated into a team decision.
    \item Using CIFAR-10 and CIFAR-100, we investigate how well a TM's classification confidence corresponds to its ability to classify images accurately. When the TM is confident, does it get higher accuracy, and when it is uncertain, does it get lower accuracy?
    \item We further provide empirical evidence that a single TM becomes a specialist rather than a generalist. As such, it obtains high accuracy on a subset of the data, decided by how we booleanize the input.
    \item To investigate whether different TM specialists can be complementary, we implement three specializations: Histogram of Gradients, Adaptive Gaussian Thresholding, and Color Thermometers.
    \item We finally evaluate the team performance on Fashion-MNIST, CIFAR-10, and CIFAR-100, reporting a percentage increase of $2$ points for Fashion-MNIST, $12$ points for CIFAR-10, and $9$ points for CIFAR-100.
\end{itemize}
In conclusion, our team-based approach sets the new state-of-the-art performance for TMs across Fashion-MNIST, CIFAR-10, and CIFAR-100.
 
\section{Tsetlin Machine Basics}

\paragraph{Input and Output.} A TM processes a vector $\mathbf{x}=[x_1,\ldots,x_o]$ of $o$ Boolean features as input, to be classified into one of $m$ classes, $y \in \{1, 2, \ldots, m\}$. 

\paragraph{Pattern Representation.} Extending the features in $\mathbf{x}$ with their negated counterparts gives the literal set $L$:
\begin{equation}
    L = \{x_1,\ldots,x_o,\neg{x}_1,\ldots,\neg{x}_o\}.
\end{equation}
A TM ANDs literals to represent sub-patterns, referred to as \emph{conjunctive clauses}. The number of clauses per class is given by a user set parameter~$n$. Half of the clauses gets positive polarity ($+$). The other half gets negative polarity ($-$). Each clause $C_j^{i,p}, i \in \{1, 2, \ldots, m\}, j \in \{1, 2, \ldots, n/2\}, p \in \{-,+\},$ then becomes:
\begin{equation}
C_j^{i,p}(\mathbf{x})=\bigwedge_{l_k \in L_j^{i,p}} l_k.
\end{equation}
\noindent Above, $i$ is the class index, $j$ is the index of the clause, $p$ its polarity, while $L_j^{i,p}$ is a subset of the literals $L$, $L_j^{i,p} \subseteq L$. For example, the clause $C_1^{2,+}(\mathbf{x}) = \neg x_1 \land x_2$ belongs to class $2$, has index $1$, polarity $+$, and consists of the literals $L_1^{2,+} = \{\neg x_1, x_2\}$. Accordingly, the clause outputs~$1$ if $x_1 =0$ and $x_2 = 1$, and $0$ otherwise.

\paragraph{Classification.} The clause outputs are combined into a classification decision by identifying the class with the largest clause sum:
\begin{equation}
\textstyle
\hat{y} = \mathrm{argmax}_{i}\left(\sum_{j=1}^{n/2} C_j^{i,+}(\mathbf{x}) - \sum_{j=1}^{n/2} C_j^{i,-}(\mathbf{x})\right).
\end{equation}
Notice how the positive clauses vote in favour of their class, while the negative clauses vote against.

\paragraph{Classification Confidence.} While there are various ways to represent classification confidence \cite{abeyrathna2020confidence}, a simple approach is to use the max class sum:
\begin{equation}
\textstyle
c_\mathrm{max} = \mathrm{max}_{i}\left(\sum_{j=1}^{n/2} C_j^{i,+}(\mathbf{x}) - \sum_{j=1}^{n/2} C_j^{i,-}(\mathbf{x})\right).\label{eqn:max_class_sum}
\end{equation}
Accordingly, $c_\mathrm{max}$ measures the confidence in classification $\hat{y}$.

\paragraph{Learning Process.} For an introduction to how TMs learn clauses from data, we refer the reader to the original TM paper~\cite{granmo2018tsetlin}. The present paper only relies on the classification procedure.

\begin{figure}[ht]
\centering
\includegraphics[width=0.9\columnwidth]{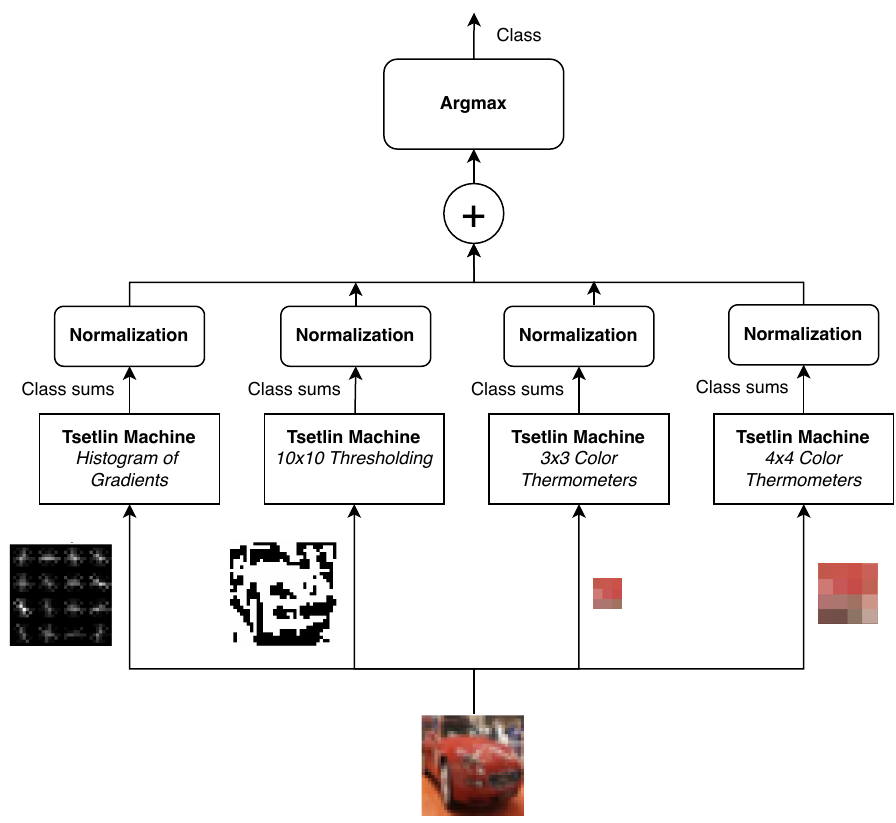}
\caption{A TM Composite of specialized TMs that enables plug-and-play collaboration.}\label{fig:plug-and-play}
\end{figure}

\section{Architecture for Plug-and-Play Collaboration Between Specialized Tsetlin Machines}

The plug-and-play architecture for TM collaboration is shown in Figure~\ref{fig:plug-and-play}. As seen, the architecture consists of multiple TMs $t$, $t \in \{1, 2, \ldots, r\}$, forming a \emph{TM Composite}. Operating alone, a TM $t$ outputs the class $i$ with the largest class sum for each input $\mathbf{x}_d \in \mathcal{X} = \{\mathbf{x}_1, \mathbf{x}_2, \ldots,  \mathbf{x}_q\}$:
\begin{equation}
\hat{y}_{t,d} = \mathrm{argmax}_{i}\left(\sum_{j=1}^{n/2} C_{t,j}^{i,+}(\mathbf{x}_d) - \sum_{j=1}^{n/2} C_{t,j}^{i,-}(\mathbf{x}_d)\right).
\end{equation}
Above, $\mathcal{X}$ is a data set of $q$ inputs and $d$ is the index of the input under consideration.

When collaborating in a composite, each TM collaborator $t$ instead outputs its class sums $c^i_{t,d}$. In brief, the class sum $c^i_{t,d}$ signifies the confidence of TM $t$ in class $i$ for input $\mathbf{x}_d$:
\begin{equation}
c^i_{t,d} = \sum_{j=1}^{n/2} C_{t,j}^{i,+}(\mathbf{x}_d) - \sum_{j=1}^{n/2} C_{t,j}^{i,-}(\mathbf{x}_d).
\end{equation}

After calculating the class sums for TM $t$, they are normalized by dividing by the difference $\alpha_t$ between the largest and smallest class sums in the input set $\mathcal{X}$:

\begin{equation}
\alpha_t = \mathrm{max}_{d,i}(c^i_{t,d}) - \mathrm{min}_{d,i}(c^i_{t,d}).
\end{equation}

The normalized class sums, in turn, are added together, forming the class sums of the TM Composite as a whole. The maximum value of these decides the class output in the final step:

\begin{equation}
\hat{y}_d = \mathrm{argmax}_{i}\left(\sum_{t=1}^{r} \frac{1}{\alpha_t} c^i_{t,d}\right).
\end{equation}

\section{Empirical Evaluation}

We first investigate how faithfully a TM assesses its classification accuracy. We further demonstrate that TMs with different booleanization strategies develop distinct specialities, laying a foundation for collaboration.  Finally, we evaluate the resulting TM Composites.

\begin{figure}[ht]
\centering
\includegraphics[width=0.75\columnwidth]{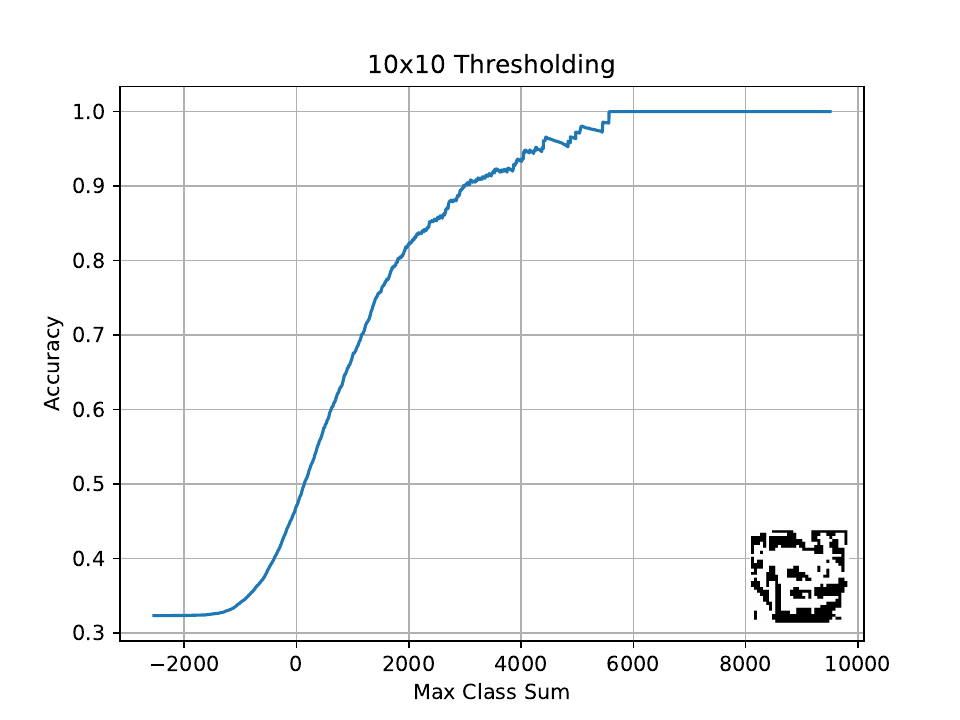}
\caption{TM accuracy (y-axis) at different confidence levels (x-axis) for CIFAR-100, employing Adaptive Gaussian Thresholding per color channel.}\label{fig:cifar100_thresholding_class_sum}
\end{figure}

\paragraph{Does the Tsetlin Machine Know When it Does Not Know?} Figure~\ref{fig:cifar100_thresholding_class_sum} relates the accuracy of a \emph{Thresholding TM} to its classification confidence on CIFAR-100. The Thresholding TM booleanizes the data using Adaptive Gaussian Thresholding\footnote{\href{https://docs.opencv.org/4.x/d7/d4d/tutorial_py_thresholding.html}{https://docs.opencv.org/4.x/d7/d4d/tutorial\_py\_thresholding.html}.} per color channel. We here use $8~000$ weighted clauses per class~\cite{abeyrathna2021integer}, a voting margin $T=2000$, specificity $s=10.0$, and a $10 \times 10$ convolution window (see \cite{granmo2019convtsetlin} for an explanation of the hyperparameters). Along the x-axis, we rank the $10~000$ test images of CIFAR-100 from lowest to highest max class sum, $c_\mathrm{max}$ in Eqn. \ref{eqn:max_class_sum}. The y-axis shows accuracy on the test images from the x-axis confidence level and upwards. When confidence is low, the TM operates at its lowest accuracy. As confidence increases, accuracy grows, eventually reaching 100\%. Accordingly, the max class sum seems to be faithful to performance. Informally, one could say that a TM \emph{knows when it does not know}, a prerequisite for collaboration.

\begin{figure}[ht]
\centering
\includegraphics[width=0.75\columnwidth]{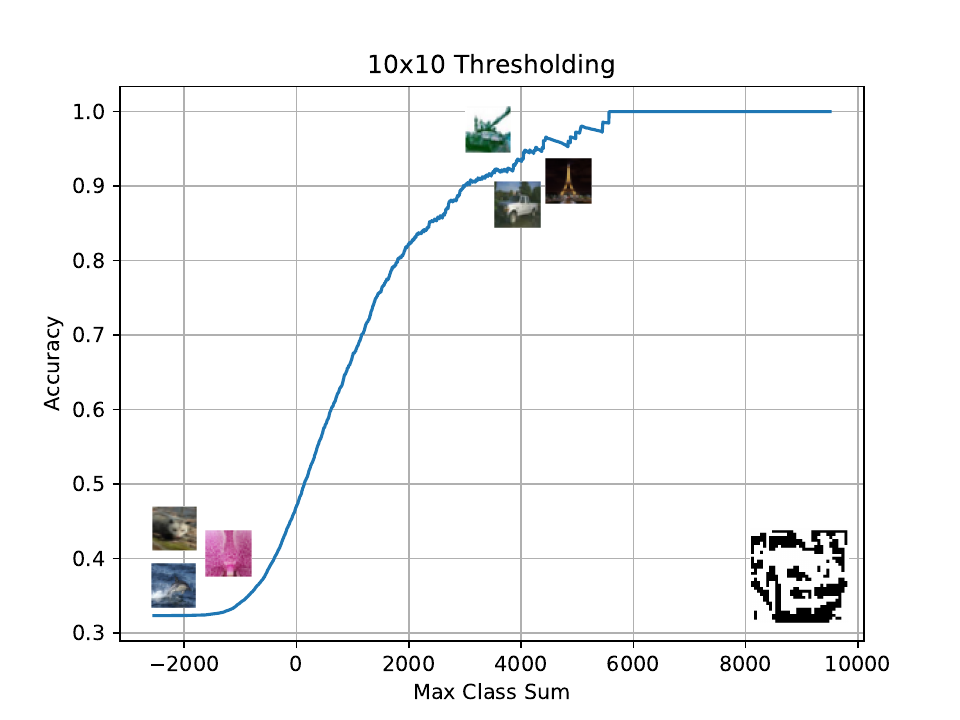}
\caption{Six images at the lower and higher ends of the confidence spectrum of the Thresholding TM.}\label{fig:cifar100_thresholding_analysis}
\end{figure}

\paragraph{Is the Tsetlin Machine a Generalist or a Specialist?} We next investigate six images at the lower and higher ends of the confidence spectrum. The pictures added to Figure~\ref{fig:cifar100_thresholding_analysis} demonstrate that the Thresholding TM has specialized in recognizing larger pixel structures, operating at high confidence and accuracy. However, for images characterized by color texture, accuracy/confidence is low. It looks like the TM has prioritized high accuracy on a subset of the data, becoming a specialist.

\begin{figure}[ht]
\centering
\includegraphics[width=0.75\columnwidth]{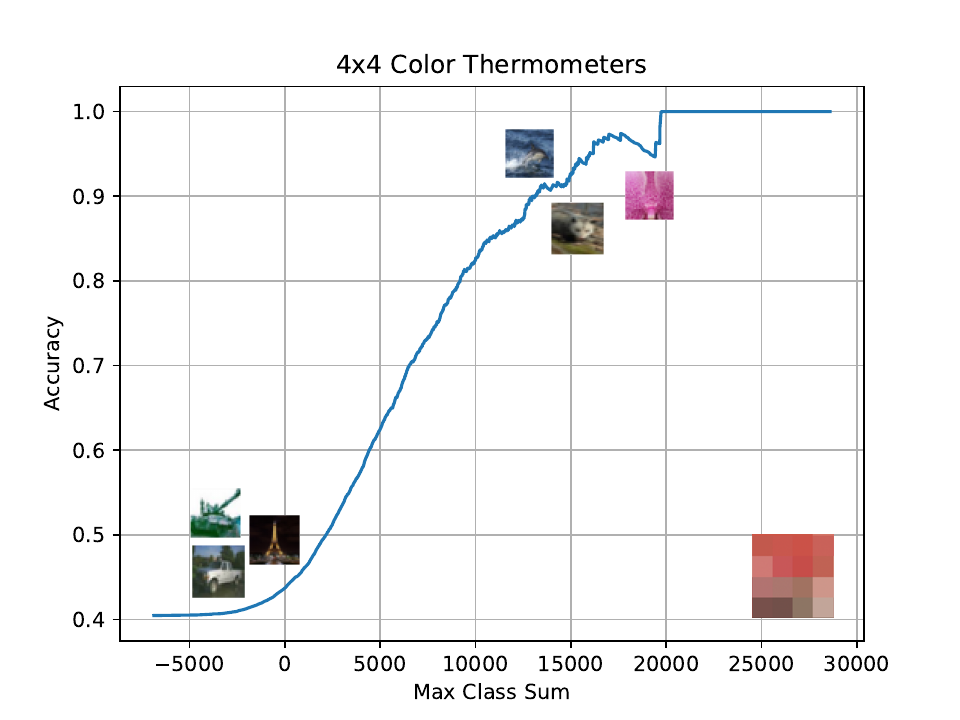}
\caption{Six images at the lower and higher ends of the confidence spectrum of the Color Thermometer TM.}\label{fig:cifar100_thermometers_analysis}
\end{figure}

\paragraph{Can Different TM Specialists be Complementary?} Now consider the confidence and accuracy of the Color Thermometer TM in Figure~\ref{fig:cifar100_thermometers_analysis}. The Color Thermometer TM booleanizes the input using 8-level thermometer encoding~\cite{buckman2018thermometer} per color channel. We use $8~000$ weighted clauses per class, a voting margin $T=6~000$, specificity $s=2.5$, and a $4 \times 4$ convolution window. Notice how the resulting TM obtains high accuracy from color texture. The weakness of the Thresholding TM seems to be the strength of the Color Thermometer TM, and vice versa.

\begin{table}[hb]
    \centering
    \begin{tabular}{c|c|c|c}
    \textbf{TM Specialist}&\textbf{CIFAR-10}&\textbf{CIFAR-100}&\textbf{Fashion-MNIST}\\
    \hline
        \multirow{3}{*}{10x10 Thresholding}&$2K$ \emph{weighted clauses}&$16K$ \emph{weighted clauses}&$8K$ \emph{weighted clauses} \\
        &$T=500$&$T=4~000$&$T=2~000$\\
        &$s=10.0$&$s=10.0$&$s=10.0$\\
        \hline
        \multirow{3}{*}{3x3/4x4 Color Thermometers}&$2K$ \emph{weighted clauses}&$16K$ \emph{weighted clauses}&$8K$ \emph{weighted clauses} \\
        &$T=1~500$&$T=12~000$&$T=6~000$\\
        &$s=2.5$&$s=2.5$&$s=2.5$\\
        \hline
        \multirow{3}{*}{Histogram of Gradients}&$2K$ \emph{clauses}&$16K$ \emph{clauses}&$8K$ \emph{clauses} \\
        &$T=50$&$T=400$&$T=200$\\
        &$s=10.0$&$s=10.0$&$s=10.0$\\
    \end{tabular}
    \caption{The TM hyperparameters of each experiment. The configurations use a budget of $32$ literals~\cite{abeyrathna2023budget}.}
    \label{tab:hyperparameters}
\end{table}

\begin{table}[hb]
    \centering
    \begin{tabular}{c|c|c|c}
    \textbf{TM Composites}&\textbf{CIFAR-10}&\textbf{CIFAR-100}&\textbf{Fashion-MNIST}\\
    \hline
         10x10 Thresholding&$57.0$&$35.1$&$91.1$ \\
         3x3 Color Thermometers&$62.1$&$42.7$&$90.5$ \\
         4x4 Color Thermometers&$62.7$&$43.4$&$90.6$\\
         Histogram of Gradients&$63.5$&$25.4$&$91.1$\\
         10x10T + HoG&$68.0$&$38.9$&$92.4$\\
         10x10T + 3x3C&$68.6$&$49.6$&$92.3$\\
         10x10T + 4x4C&$68.7$&$49.2$&$92.2$\\
         10x10T + 3x3C + 4x4C&$71.7$&$50.3$&$92.4$\\
         10x10T + 3x3C + 4x4C + HoG&$75.1$&$52.2$&$93.0$
    \end{tabular}
    \caption{Accuracy after 100 epochs.}
    \label{tab:accuracy}
\end{table}

\begin{figure}[!!ht]
\centering
\includegraphics[width=0.75\columnwidth]{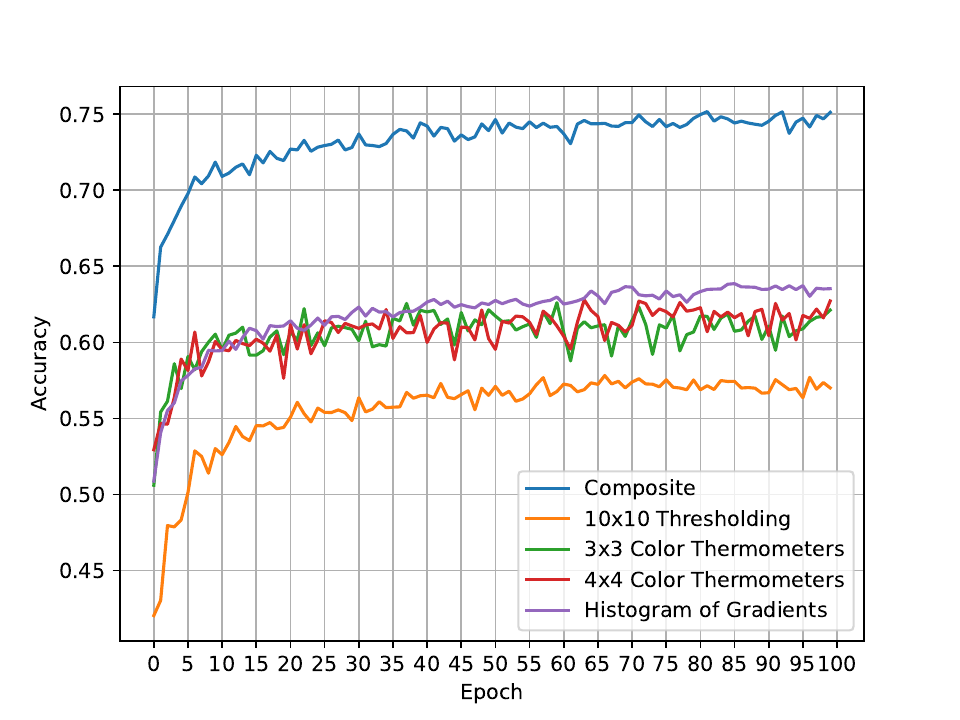}
\caption{CIFAR-10 accuracy, epoch-by-epoch.}\label{fig:cifar10_epochs}
\end{figure}

\begin{figure}[!!ht]
\centering
\includegraphics[width=0.75\columnwidth]{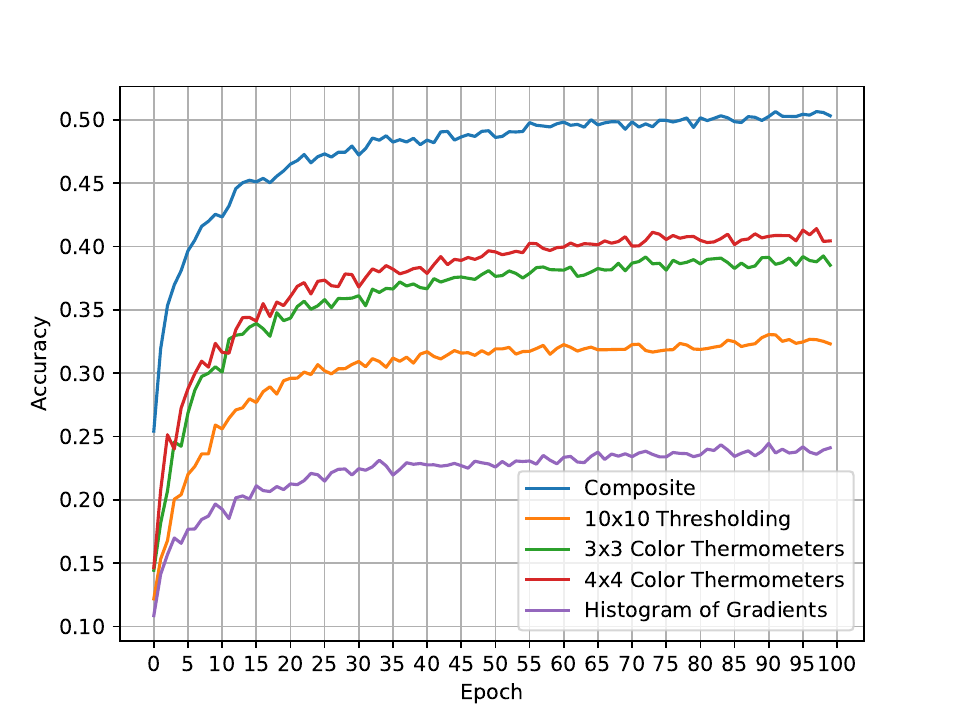}
\caption{CIFAR-100 accuracy, epoch-by-epoch.}\label{fig:cifar100_epochs}
\end{figure}

\paragraph{Can Specialist Tsetlin Machines Collaborate Successfully?} We are now ready to deploy our plug-and-play architecture for creating TM Composites. First, we introduce another TM specialist based on a Histogram of Gradients.\footnote{The Histogram of Gradients TM reuses the setup from \href{https://github.com/hudara/cifar-10}{https://github.com/hudara/cifar-10}.} Table \ref{tab:hyperparameters} contains the hyperparameters of each TM specialist for Fashion-MNIST, CIFAR-10, and CIFAR-100. Now, consider the accuracy of the various TM Composites in Table \ref{tab:accuracy} and observe how additional TM specialists improve accuracy. Indeed, the full-blown composite increases accuracy on Fashion-MNIST by two percentage points, CIFAR-10 by twelve points, and CIFAR-100 by nine points, yielding new state-of-the-art results for TMs. Finally, consider accuracy epoch-by-epoch for CIFAR-10 in Figure~\ref{fig:cifar10_epochs} and CIFAR-100 in Figure~\ref{fig:cifar100_epochs}. The composite performance is superior from the first epoch and stays ahead throughout the learning process.

\section{Conclusions and Further Research}

In this paper, we introduced the concept of \emph{TM Composites}. A TM Composite merges multiple TMs with different specialities, significantly boosting team performance. The architecture is plug-and-play in that the individual TMs can train independently and then combine in any way, at any time, without fine-tuning. The architecture yields new state-of-the-art results for TMs on Fashion-MNIST, CIFAR-10, and CIFAR-100, using three image processing methods for demonstration.

The TM Composites raise several research questions:
\begin{itemize}
\item What other specializations (image processing techniques) can boost TM Composites further?
\item Can we design a light optimization layer that enhances the collaboration accuracy, e.g., by weighting the specialists based on their performance?
\item Are there other ways to normalize and integrate the perspective of each TM?
\item Can we find a way to fine-tune the TM specialists to further support collaboration?
\item What is the best approach to organizing a library of composable pre-trained TMs?
\item How can we compose the most efficient team with a given size?
\item What is the best strategy for decomposing a complex feature space among independent TMs?
\item Does our approach extend to other tasks beyond image classification?
\end{itemize}
Overall, we envision that TM Composites will enable the TM to become an ultra-low energy and transparent alternative to state-of-the-art deep learning approaches on more tasks and datasets.

\bibliographystyle{abbrv}
\bibliography{References}

\begin{thebibliography}{10}

\bibitem{abeyrathna2023budget}
K.~D. Abeyrathna, A.~A.~O. Abouzeid, B.~Bhattarai, C.~Giri, S.~Glimsdal,
  O.~Granmo, L.~Jiao, R.~Saha, J.~Sharma, S.~A. Tunheim, and X.~Zhang.
\newblock {Building Concise Logical Patterns by Constraining Tsetlin Machine
  Clause Size}.
\newblock In {\em Proceedings of the Thirty-Second International Joint
  Conference on Artificial Intelligence, {IJCAI} 2023, 19th-25th August 2023,
  Macao, SAR, China}, pages 3395--3403. ijcai.org, 2023.

\bibitem{abeyrathna2020confidence}
K.~D. Abeyrathna, O.-C. Granmo, and M.~Goodwin.
\newblock {On Obtaining Classification Confidence, Ranked Predictions and AUC
  with Tsetlin Machines}.
\newblock In {\em 2020 IEEE Symposium Series on Computational Intelligence
  (SSCI)}, pages 662--669, 2020.

\bibitem{abeyrathna2021integer}
K.~D. Abeyrathna, O.-C. Granmo, and M.~Goodwin.
\newblock {Extending the Tsetlin Machine With Integer-Weighted Clauses for
  Increased Interpretability}.
\newblock {\em IEEE Access}, 9:8233--8248, 2021.

\bibitem{sauer2021counterfactual}
A.~G. Axel~Sauer.
\newblock {Counterfactual Generative Networks}.
\newblock In {\em International Conference on Learning Representations (ICLR)},
  2021.

\bibitem{Bender2021Parrots}
E.~M. Bender, T.~Gebru, A.~McMillan-Major, and S.~Shmitchell.
\newblock {On the Dangers of Stochastic Parrots: Can Language Models Be Too
  Big?}
\newblock In {\em FAccT '21: Proceedings of the 2021 ACM Conference on
  Fairness, Accountability, and Transparency}, FAccT '21, page 610–623, New
  York, NY, USA, 2021. Association for Computing Machinery.

\bibitem{buckman2018thermometer}
J.~Buckman, A.~Roy, C.~Raffel, and I.~J. Goodfellow.
\newblock {Thermometer Encoding: One Hot Way To Resist Adversarial Examples}.
\newblock In {\em 6th International Conference on Learning Representations,
  {ICLR} 2018, Vancouver, BC, Canada, April 30 - May 3, 2018, Conference Track
  Proceedings}. OpenReview.net, 2018.

\bibitem{granmo2018tsetlin}
O.-C. {Granmo}.
\newblock {The Tsetlin Machine - A Game Theoretic Bandit Driven Approach to
  Optimal Pattern Recognition with Propositional Logic}.
\newblock {\em arXiv preprint arXiv:1804.01508}, 2018.

\bibitem{granmo2019convtsetlin}
O.-C. {Granmo}, S.~{Glimsdal}, L.~{Jiao}, M.~{Goodwin}, C.~W. {Omlin}, and
  G.~T. {Berge}.
\newblock {The Convolutional {T}setlin Machine}.
\newblock {\em arXiv preprint arXiv:1905.09688}, 2019.

\bibitem{lucas1995}
P.~J.~F. Lucas.
\newblock {Logic engineering in medicine}.
\newblock {\em The Knowledge Engineering Review}, 10(2):153–179, 1995.

\bibitem{maheshwari2023}
S.~Maheshwari, T.~Rahman, R.~Shafik, A.~Yakovlev, A.~Rafiev, L.~Jiao, and O.-C.
  Granmo.
\newblock {REDRESS: Generating Compressed Models for Edge Inference Using
  Tsetlin Machines}.
\newblock {\em IEEE Transactions on Pattern Analysis and Machine Intelligence},
  45(9):11152--11168, 2023.

\bibitem{Rudin2019}
C.~Rudin.
\newblock {Stop explaining black box machine learning models for high stakes
  decisions and use interpretable models instead}.
\newblock {\em Nature Machine Intelligence}, 1(5):206--215, 2019.

\bibitem{Saha2022}
R.~Saha, O.-C. Granmo, V.~Zadorozhny, and M.~Goodwin.
\newblock {A relational Tsetlin machine with applications to natural language
  understanding}.
\newblock {\em Journal of Intelligent Information Systems}, 2022.

\bibitem{Schwartz2020GreenA}
R.~Schwartz, J.~Dodge, N.~Smith, and O.~Etzioni.
\newblock {Green AI}.
\newblock {\em Communications of the ACM}, 63:54 -- 63, 2020.

\bibitem{scholkopf2021causal}
B.~Schölkopf, F.~Locatello, S.~Bauer, N.~R. Ke, N.~Kalchbrenner, A.~Goyal, and
  Y.~Bengio.
\newblock {Towards Causal Representation Learning}, 2021.

\bibitem{dropclause}
J.~Sharma, R.~Yadav, O.-C. Granmo, and L.~Jiao.
\newblock {Drop Clause: Enhancing Performance, Robustness and Pattern
  Recognition Capabilities of the Tsetlin Machine}.
\newblock {\em Proceedings of the AAAI Conference on Artificial Intelligence},
  37(11):13547--13555, Jun. 2023.

\bibitem{Tsetlin1961}
M.~L. Tsetlin.
\newblock {{On behaviour of finite automata in random medium}}.
\newblock {\em Avtomat. i Telemekh}, 22(10):1345--1354, 1961.

\end{thebibliography}

\end{document}